\documentclass[11pt]{article}
\usepackage{acl2015}
\usepackage{times}
\usepackage{url}
\usepackage{latexsym}
\usepackage{amsmath}
\usepackage{float} 
\usepackage{graphicx} 
\usepackage{amssymb}
\usepackage{array}
\usepackage{subcaption}
\usepackage{algorithm}
\usepackage{algpseudocode} 
\usepackage{amsmath}

\title{Optimizing Vision Transformers with Data-Free Knowledge Transfer}

\author{Gousia Habib 1*\\
   Postdoc fellow \\ IIT Delhi\\  \And
  Damandeep Singh \\
   M.Tech Student\\  IIT Delhi\\
  \And
  Ishfaq Ahmad Malik \\
    Assistant Professor\\ Shoolini University 
  \And
  Brejesh Lall \\
Professor \\  IIT Delhi   \\
  \\   }

\date{}

\begin{document}
\maketitle
\begin{abstract}
\textit{The groundbreaking performance of transformers in Natural Language Processing (NLP) tasks has led to their replacement of traditional Convolutional Neural Networks (CNNs), owing to the efficiency and accuracy achieved through the self-attention mechanism. This success has inspired researchers to explore the use of transformers in computer vision tasks to attain enhanced long-term semantic awareness. Vision transformers (ViTs) have excelled in various computer vision tasks due to their superior ability to capture long-distance dependencies using the self-attention mechanism. Contemporary ViTs like  Data Efficient Transformers (DeiT) can effectively learn both global semantic information and local texture information from images, achieving performance comparable to traditional CNNs. However, their impressive performance comes with a high computational cost due to very large number of parameters, hindering their deployment on devices with limited resources like smartphones, cameras, drones etc. Additionally, ViTs require a large amount of data for training to achieve performance comparable to benchmark CNN models. Therefore, we identified two key challenges in deploying ViTs on smaller form factor devices: the high computational requirements of large models and the need for extensive training data. As a solution to these challenges, we propose compressing large ViT models using Knowledge Distillation (KD), which is implemented data-free to circumvent limitations related to data availability. Additionally, we conducted experiments on object detection within the same environment in addition to classification tasks. Based on our analysis, we found that data-free knowledge distillation is an effective method to overcome both issues, enabling the deployment of ViTs on less resource-constrained devices.}
\end{abstract}

\section{Introduction}
At the vanguard of computer vision advances CNNs  have produced state-of-the-art outcomes in a range of tasks, such as image classification [1], object detection [2], image segmentation [3], facial recognition [4], and scene interpretation [5]. Even with their achievements, CNNs have several drawbacks when used for jobs involving vision. Through their localized receptive fields and weight-sharing mechanisms, CNNs have great success capturing local features in images, but they struggle to model global context and long-range dependencies. Their intrinsic design, which prioritises local connectivity via convolutional layers, gives rise to this constraint. Second, CNNs limited capacity to generalise across different dimensions and contexts within an image might be attributed to their fixed receptive fields, which are defined by the size of the convolutional kernels and the overall network architecture.Last but not least, the considerable number of parameters required, especially in deeper CNN architectures [6], leads to high computational costs, making them unsuitable for applications requiring real-time computation.\\
Researchers have been looking into ViTs suitability for CV tasks in recent years due to their remarkable performance in NLP. In this sense, Alexey Dosovitskiy et al.'s introduction of  ViTs [7] represents a major paradigm change. ViTs break down an image into patches and feed a transformer model a series of linear embeddings of these patches. By treating picture patches similarly to tokens (words) in NLP applications, this method improves the model's ability to capture global context and long-range dependencies.\\
ViTs are more semantically aware than conventional CNN architectures because they use self-attention methods to capture global context and  long range dependencies between image tokens. These benefits do, however, come with several difficulties that must be overcome in order to create ViT based models as a  deployable applications for the real world.\\
The fact that ViTs can have a lot of parameters makes them computationally expensive and resource-intensive, which poses a critical challenge, especially when deploying them on edge devices with constrained resources. Additionally, in order for ViTs to function on par with benchmark CNN models, a substantial amount of training data is needed.\\
Compressing ViTs models into smaller versions that are feasible to be deployed on  edge devices with limited memory and processing power is crucial to combat these challenges. This is accomplished by transferring knowledge from a bigger pre-trained instructor model to a smaller student model using technique known as KD [8]. which enables deployment on devices with limited resources. Using the original training data is a potential technique for this compression, albeit it may not always be available due to transmission limits or privacy considerations. Under such circumstances, using artificial intelligence data becomes a viable option.CNNs can be effectively compressed using this method [9].\\
However, due to the significant differences in model structure and computational mechanisms between CNNs and ViTs, it remains an open question whether a similar paradigm is suitable for ViTs.\\
This work presents a unique technique for compressing ViTs utilizing synthetic data, specifically designed for object classification and object detection.
\section{Related Work}
The transformer architecture was introduced by Vaswani et al. [7] in 2017, revolutionizing NLP by capturing long-range dependencies within textual space. Transformers model interactions between all words in a sentence simultaneously, as opposed to Recurrent Neural Networks (RNNs) [10], which process word relationships sequentially. A paradigm shift improved tasks such as answering questions, summarizing information, and translating it. 
Taking advantage of this transformative approach, ViTs emerged as  a groundbreaking approach in CV, utilizing the attention based architecture developed for NLP to process image data. ViTs have defied the conventional supremacy of CNNs by demonstrating impressive performance on a variety of image recognition benchmarks since their introduction by Dosovitskiy et al. in paper [11] in 2020. The success of transformers in NLP led researchers to study the applicability of transformers to CV. In spite of the fact that CNNs are effective, they are limited in terms of their ability to capture global context because of their narrow receptive fields.\\
A fundamental aspect of the transformer's function is its ability to model relationships throughout an input sequence through the use of self-attention [12]. As part of ViTs, self-attention enables the model to evaluate the relative importance of individual visual patches, aiding in comprehensive image understanding. As a result of self-attention, attention maps [13] are generated for each patch, which visually represent the model's focus.The maps provide insight into how the model processes and prioritizes different areas of an image.\\
The transformer architecture has made significant progress across a range of state-of-the-art applications [14]. In spite of their successes, transformers today are dependent on self-attention mechanisms that have a quadratic time and space complexity as input length increases [15]. There are several ways to accelerate self-attention mechanisms [16] to achieve sub-quadratic running times, but most of these approaches lack rigorous error guarantees.\\
Furthermore, ViTs require extensive training datasets to establish their inductive biases. As a result, these algorithms can't be applied to real-time, resource-constrained devices because of this requirement [17, 18].\\
A promising solution to these limitations lies in KD [19], where a smaller, simpler model (the student) is trained to mimic the behaviour of a bigger, more complex model (the teacher) in a process known as KD. KD can help mitigate some of the resource constraints associated with Vision Transformers (ViTs) in real-time applications by compressing models in real-time [20].\\
This methodology is especially useful for deploying models without appreciably sacrificing performance on devices with constrained computational capabilities, like embedded systems or mobile phones. The ``knowledge'' provided by the teacher model, which is usually a very accurate but computationally costly model, takes the form of output probabilities also known by the name as soft logits [21]. Compared to the hard labels (ground truth) [22], these probabilities, sometimes referred to as soft targets [21], provide more detailed information about the uncertainty and correlations between various classes. The learner can improve their ability to generalise by training their model on these soft targets. This will allow them to achieve great accuracy with a comparatively small number of parameters [23].\\
    However, the traditional KD model [24] assumes that the student model has access to all or part of the teacher's training data.
    The original training data must, however, be restricted when used in real-world applications. In cases involving privacy-sensitive medical data, which may contain personal information or proprietary data, this issue becomes particularly relevant.Therefore, conventional KD methods[24, 25] are no longer sufficient to address the challenges faced in these contexts [26].\\
    A compelling alternative to these limitations is the Data-Free Knowledge Distillation (DFKD) protocol [27, 28, 29]. To facilitate the transfer of knowledge from a pre-trained teacher model to a student model, DFKD generates synthetic images without access to original training data. In this method, the student attempts to match the teacher's predictions on synthetic data, while a generator creates samples meant to mislead the student, aligning with the teacher's confidence. By utilizing an adversarial framework, it is possible to explore synthetic distributions and transfer knowledge between models while maintaining data privacy [30, 31].\\
Even though DFKD has demonstrated promising results, several challenges still remain unanswered. There is the potential for discrepancy between synthetic data and original data distributions, which can introduce bias into the student model's learning process. Student networks may exhibit bias as a result of noise present in synthetic images that distorts their focus and learning regions. In addition, the frequently used Kullback-Leibler (KL) [32, 33, 34] divergence constraint between student and teacher networks in existing DFKD methods may perform less well with synthetic data, leading to reduced knowledge transfer accuracy.\\
In order to address the challenges associated with DFKD, we have developed a novel approach that combines adversarial learning with transformers and employs data-free distillation with a custom loss function. Our approach entails using transformers in an adversarial learning framework to generate high-quality synthetic samples similar to the original data. We also use data-free distillation, in which knowledge is transferred from the teacher to the student model without reusing the original data.\\
Our approach optimizes this process by applying an attention class loss function that aligns the student model's attention mechanisms with the teacher's. As a result of this integrated approach, issues such as distribution mismatch and bias are mitigated, thereby enhancing knowledge transfer effectiveness.
\section{Requirement for Data free Distillation.}
KD is crucial for compressing large, pre-trained models into smaller, more efficient versions while retaining much of the original model's performance. In the context of  ViTs, this becomes particularly important given their substantial size and computational demands. However, traditional knowledge distillation methods rely on access to the original training data, which may not always be feasible due to several reasons:
\begin{itemize}
    \item \textbf{Privacy Concerns}: In many applications, especially those involving sensitive information such as medical images or personal data, privacy regulations (e.g., GDPR, HIPAA) restrict access to the original datasets. Sharing or using these datasets for further training can lead to privacy violations and legal issues.
    \item \textbf{Data Availability Issues}: Sometimes, the original training data may no longer be available. This can happen due to data deletion policies, data corruption, or the data being owned by third parties who are unwilling or unable to share it.
    \item \textbf{Transmission Restrictions}: In scenarios where data needs to be transmitted across different geographical locations or organizations, there may be bandwidth limitations or regulatory restrictions that prevent the transfer of large datasets. This is particularly relevant in distributed and federated learning settings where data privacy and sovereignty are of utmost importance.
    \end{itemize}
Given these challenges, there is a critical  need to develop data-free distillation methods that can effectively transfer knowledge from a large teacher model to a smaller student model without requiring the original dataset. \\
Mathematically, Data-Free Knowledge Distillation (DFKD) can be formulated as follows:

If \( D = \{X \in \mathbb{R}^{c \times h \times w}, Y = 1, 2, \ldots, K\} \) gives the training dataset and labels.

and \( T(x; \theta_T) \) is a pre-trained teacher network on \( D \).

The main task for student is to minimize the losses ie:
\[
\min_{\theta_S} \mathcal{L}_{cls} + \mathcal{L}_{KL}
\]

In DFKD we learn a lightweight  classification network \( S(x; \theta_S) \) that can imitate the classification capability of \( T(x; \theta_T) \) without using \( D \).

The primary requirements for data-free distillation in ViTs include:

\begin{itemize}
    \item \textbf{Model Compression}: The distilled model should be significantly smaller and less resource-intensive than the original model while maintaining comparable performance.
    \item \textbf{Synthetic Data Generation}: Since the original data is not available, synthetic data generation techniques must be employed to create a substitute dataset that can be used for distillation.
    \item \textbf{Preservation of Knowledge}: The distilled model should preserve the essential knowledge and features learned by the larger model, ensuring that it performs well on the target tasks.
    \item \textbf{Efficiency}: The distillation process should be computationally efficient, making it feasible to run on devices with limited resources
    \item \textbf{Robustness}: The method should be robust to variations in synthetic data quality and capable of producing reliable results across different tasks and datasets.
\end{itemize}

\section{Problem Statement}

Our literature review led us to the following conclusions:\\
(a) The significant computational and data requirements of ViTs frequently impede their deployment in real-world applications; and\\
(b) Traditional KD techniques heavily rely on large datasets that may be unavailable owing to transmission limitations or privacy concerns.\\
The crucial problem of carrying out KD in ViTs without having access to the original training data is addressed in this paper. In order to bridge the gap between cutting-edge performance and real-world application, it is intended to create efficient data-free distillation techniques that enable the compression of ViTs into more manageable, compact models that can be used on devices with constrained memory and processing power. The issue can be further broken down into the subsequent smaller issues.
\begin{itemize}
    \item Adopt suitable data synthesis  technique to generate synthetic data more closer to true distribution of original data.
    \item Perform DFKD in Vision Transformers  for Classification Tasks.
    \item Perform DFKD in Vision Transformers for  Detection Tasks
    \item Transform the Outcomes as deployable model suitable  for real word application on a edge computing.
\end{itemize}
Based upon the study of issues mentioned above. The major contributions of Our research and implementation included the following novel contributions:
\begin{itemize}
    \item \textbf{Modified GAN with Patch-Level Attention}: By adding attention mechanisms at the patch level, we improved the performance of conventional GANs and increased their ability to produce high-quality images that work well with transformer-based models.
    \item \textbf{Patch Loss for Distillation}:Using the attention probe, we made it easier to distil knowledge by introducing a patch loss function. The key characteristics and representations required for effective distillation were successfully captured by this innovative loss function.
\end{itemize}
\section{Proposed Synthetic Data Generation for Data-Free Knowledge Distillation}
\label{sect:pdf}
Developing an effective strategy for creating synthetic data is one of the main obstacles to accomplishing effective data free KD without compromising performance. As an alternative to DFKD, Jiahao Wang [35] suggested using vast unlabeled data in the wild in their study ``Attention Probe: ViT in the Wild'' [35].\\
However, there are several drawbacks to this approach: (a) The unlabeled nature of wild data means that it lacks ground truth for training the student model. (b) Incorporating unwanted data from the wild could cause the student model to learn false information. (c) The availability of wild data could be restricted or non-existent for specific applications, thereby limiting its applicability. As a result, our method creates synthetic data using Transformer augmented GANs.
\subsection{Novel- Transformer augmented GANs}
\label{ssec:layout} Generative Adversarial Networks (GANs) [36] have achieved remarkable success in generating high-quality images. However, traditional GANs often require extensive training time and resources to reach desired performance levels. To address these challenges, we propose a novel approach that integrates transformers into the GAN framework, leveraging the self-attention mechanism of transformers to enhance image generation efficiency and quality.

\begin{algorithm}
\caption{Data-Free Knowledge Distillation for Vision Transformers}
\begin{algorithmic}[1]
\State \textbf{Input}: Pre-trained  Heavy teacher model $T$, Lightweight student model $S$, synthetic data generator $G$, number of epochs $E$, loss functions $\mathcal{L}_{KD}$, $\mathcal{L}_{CE}$, $\mathcal{L}_{patch}$, and a set of hyperparameters $\lambda_{KD}$, $\lambda_{CE}$, $\lambda_{patch}$.
\State \textbf{Output}: Lightweight Trained student model $S$.
\State \textbf{Initialize:} Set the parameters of student model $S$ to random values.
\State Generate synthetic data $\hat{D}$ using $G$.
\For {each epoch $e$ from 1 to $E$}
    \State \textbf{Generate Synthetic Data:}
    \For {each sample $x$ in $\hat{D}$}
        \State Create synthetic images using GANs augmented with transformers.
    \EndFor
    \State \textbf{Knowledge Distillation Loss Calculation:}
    \For {each synthetic data sample $x$}
        \State Forward pass through teacher model $T$ to obtain logits $T(x)$.
        \State Forward pass through student model $S$ to obtain logits $S(x)$.
        \State Calculate distillation loss $\mathcal{L}_{KD}$ using Kullback-Leibler (KL) Divergence between $S(x)$ and $T(x)$.
    \EndFor 
    \State \textbf{Classification Loss Calculation:}
    \For {each synthetic data sample $x$}
        \State Calculate classification loss $\mathcal{L}_{CE}$ between student model predictions $S(x)$ and the true labels.
    \EndFor
    \State \textbf{Patch Attention Loss Calculation:}
    \For  {each synthetic data sample $x$}
        \State Extract attention maps from both teacher $T$ and student $S$ models.
        \State Calculate patch attention loss $\mathcal{L}_{patch}$ to align attention maps between $T$ and $S$.
    \EndFor
    \State \textbf{Total Loss Calculation:}
    \State Combine the losses: $\mathcal{L}_{total} = \lambda_{KD} \mathcal{L}_{KD} + \lambda_{CE} \mathcal{L}_{CE} + \lambda_{patch} \mathcal{L}_{patch}.$
    \State \textbf{Backpropagation and Optimization:}
    \State Compute gradients of $\mathcal{L}_{total}$ with respect to student model $S$ parameters.
    \State Update $S$ parameters using an optimizer (e.g., Adam).
    \State \textbf{Validation:}
    \State Evaluate the performance of $S$ on a validation set using standard metrics.
\EndFor

\Return $S$
\end{algorithmic}
\end{algorithm}

\begin{figure}[H]
  \centering
\includegraphics[scale=0.35]{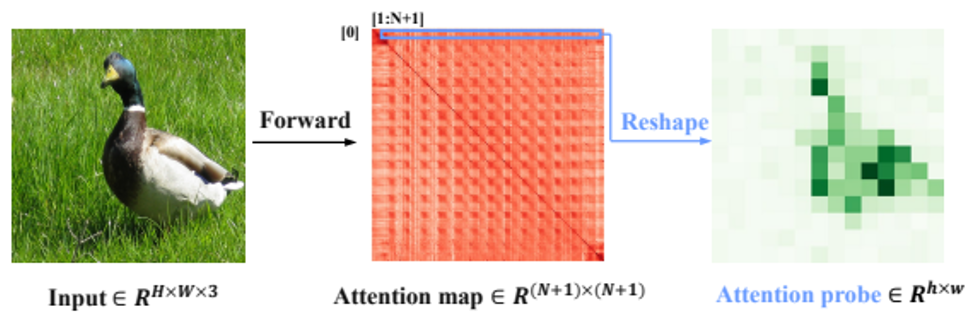} 
  \caption{Image, Attention Map and Corresponding Attention Probes}
  \label{fig:1}
\end{figure}

\textbf{The Need for Transformers in GANs}
Transformers, originally designed for NLP, have demonstrated exceptional capabilities in capturing long-range dependencies and contextual information through self-attention mechanisms. By incorporating transformers into GANs, we aim to exploit these advantages to improve the quality and diversity of generated images while reducing training time. The self-attention mechanism enables the model to focus on relevant parts of the image, thereby enhancing the generation process.\\
\textbf{Attention Probe} [35] To effectively integrate transformers into GANs, we utilize attention probes and class attention probes. An attention probe refers to the first row of the attention map from a transformer, which captures the attention distribution over the image tokens. For a generated image, the attention probe can be represented as:
\begin{equation}
AP_{\text{gen}} = A_{\text{gen}}[1:N+1]
\end{equation}
where $A_{\text{gen}}$  is the attention map of the generated image, and $AP_{\text{gen}}$ is the attention probe for the generated image figure \ref{fig:1}. 

\begin{figure}[H]
  \centering
\includegraphics[scale=0.25]{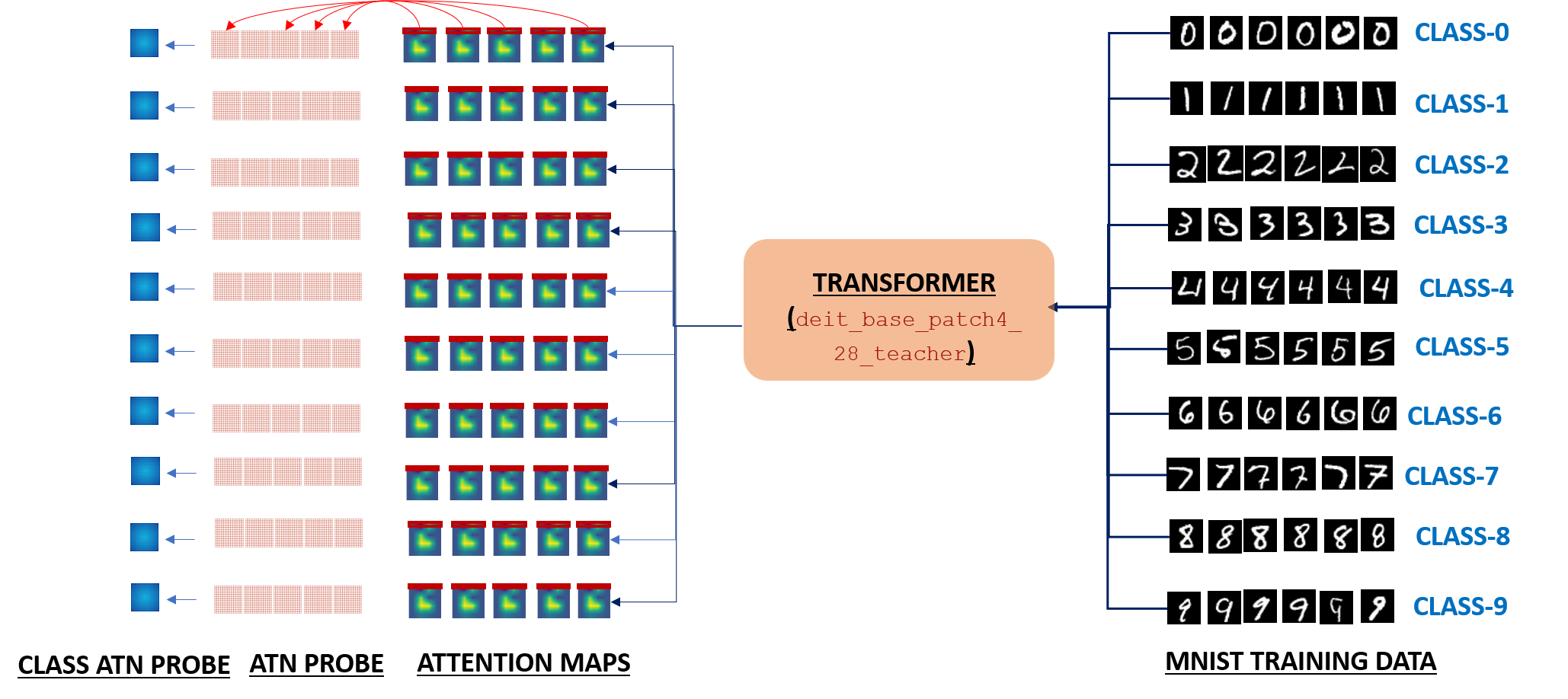}
  \caption{Class Attention Probe}
  \label{fig:2}
\end{figure}

\textbf{A class attention probe}, on the other hand, is the average of the attention probes of all training images in a particular class. This captures the typical attention distribution for that class, providing a benchmark for comparison. The class attention probe can be formulated as:
\begin{equation}
CAP_{\text{class}} = \frac{1}{K} \sum_{i=1}^{N} AP_i
\end{equation}
where $K$ is the number of training images in the class, and $AP_i$ is the attention probe of the i-th training image. Figure \ref{fig:2}  shows the class attention probe calculation. 

\begin{figure}[H]
     \centering
     \includegraphics[scale=0.30]{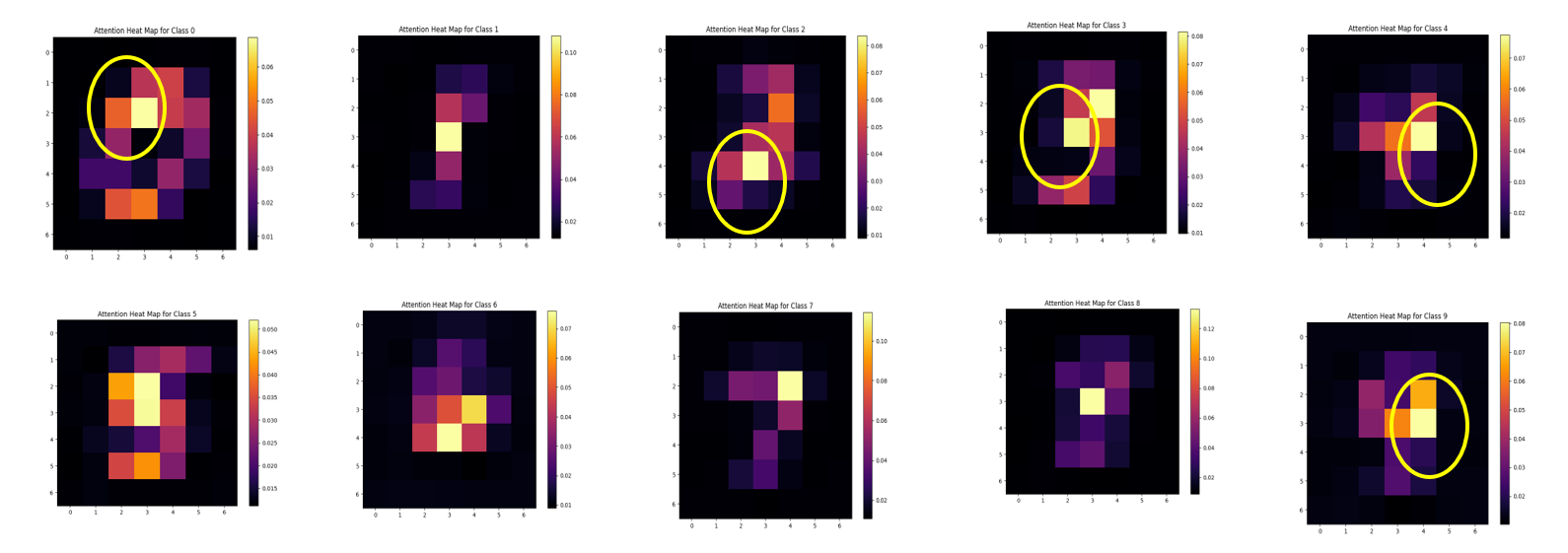}
     \caption{Class Attention Probe-MNIST}
     \label{fig: 3-1}
 \end{figure}

\begin{figure}[H]
     \centering
     \includegraphics[scale=0.30]{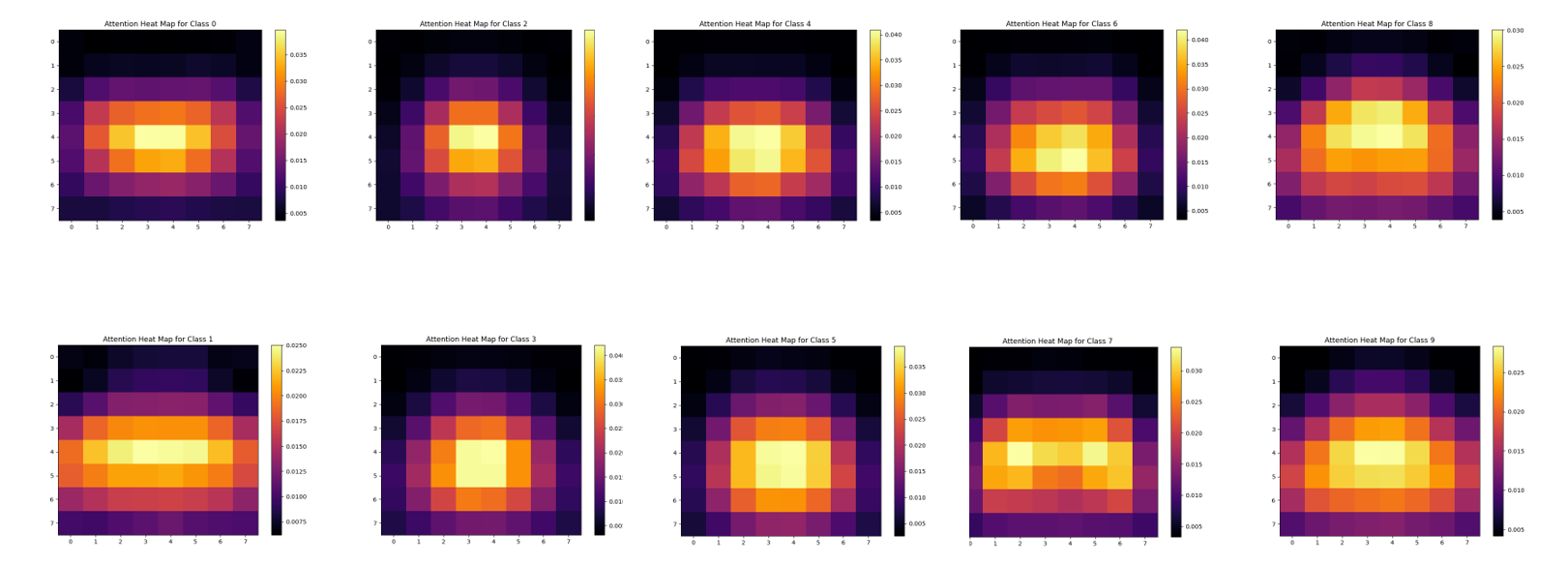}
     \caption{Class Attention Probe-CIFAR10}
     \label{fig: 3-2}
 \end{figure}

As previously noted, the attention probe is the first row in the attention map, indicating how much attention the class token allocates to each patch of the image. When examining the class attention probes for the MNIST dataset [37] as in Figure \ref{fig: 3-1} and Cifar-10 [38] as in Figure \ref{fig: 3-2}, we observe that the model focuses predominantly on patches where lines intersect within a digit Figure \ref{fig: 3-1}.  

This observation suggests that the class attention probe effectively represents the average attention distribution across patches, highlighting the regions where the model pays the most attention. By leveraging this knowledge, the generator is guided  to produce images that align with these attention patterns, ensuring the generated images are consistent with the characteristics deemed important by the model.
\subsection{Proposed Transformer-Augmented GAN Framework}
In our augmented GAN framework, we use a trained transformer to analyze the generated images. Specifically, we compare the attention probe of the generated image with the class attention probe using cosine similarity given in equation 3. This comparison helps ensure that the generated images align well with the typical attention patterns of the desired class. The loss function for the generator in our augmented GAN includes two components: the traditional adversarial loss and the attention consistency loss. Our proposed technique is illustrated in Figure \ref{fig:5a}. 
The adversarial loss ensures the generated images are realistic, while the attention consistency loss ensures that the generated images adhere to the expected attention distribution of the target class. The cosine similarity and attention consistency loss can be defined as:
\begin{equation}
\text{cosine\_similarity}(\mathbf{u}, \mathbf{v}) = \frac{\mathbf{u} \cdot \mathbf{v}}{\|\mathbf{u}\| \|\mathbf{v}\|}
\end{equation}
  Where u and v are the vectors to be compared.
\begin{equation}
L_{\text{attention}} = 1- \cos(AP_{\text{gen}}, CAP_{\text{class}})
\end{equation}
where $\cos(AP_{\text{gen}}, CAP_{\text{class}})$ represents the cosine similarity between the attention probe of the generated image and the class attention probe.\\
The overall loss function for the generator is then given by:
\begin{equation}
    L_G = L_{\text{adv}} + \lambda L_{\text{attention}}
\end{equation}
 where $L_{\text{adv}}$ is the traditional adversarial loss, $ \lambda L_{\text{attention}}$ is the attention consistency loss, and $\lambda$ is a hyperparameter that balances the two components.\\
By integrating these components, our transformer-augmented GAN framework leverages the strengths of both GANs and transformers, achieving high-quality image generation with reduced training cycles. This approach not only enhances the fidelity of the generated images but also ensures they are contextually consistent with the target class.
\subsection{Experiment Setup 1} 
To evaluate the performance of GANs, various metrics are commonly used, including Inception Score (IS) [39], Frechet Inception Distance (FID) [40], and precision-recall curves. Among these, FID is widely regarded as a robust measure because it compares the distribution of generated images to the distribution of real images, providing a comprehensive assessment of both the quality and diversity of generated images.\\
We choose FID as our primary evaluation metric because it effectively captures the similarities between real and generated images, taking into account both the mean and covariance of the features extracted by a pre-trained Inception network. This makes FID a reliable metric for comparing the performance of different GAN architectures.\\
The FID is a metric used to evaluate the quality of generated images by comparing the distributions of real and generated images in the feature space of a pretrained Inception network. The FID score is calculated as represented by equation \ref{E5}:
 \begin{figure}[H]
  \centering
 \includegraphics[scale=0.35]{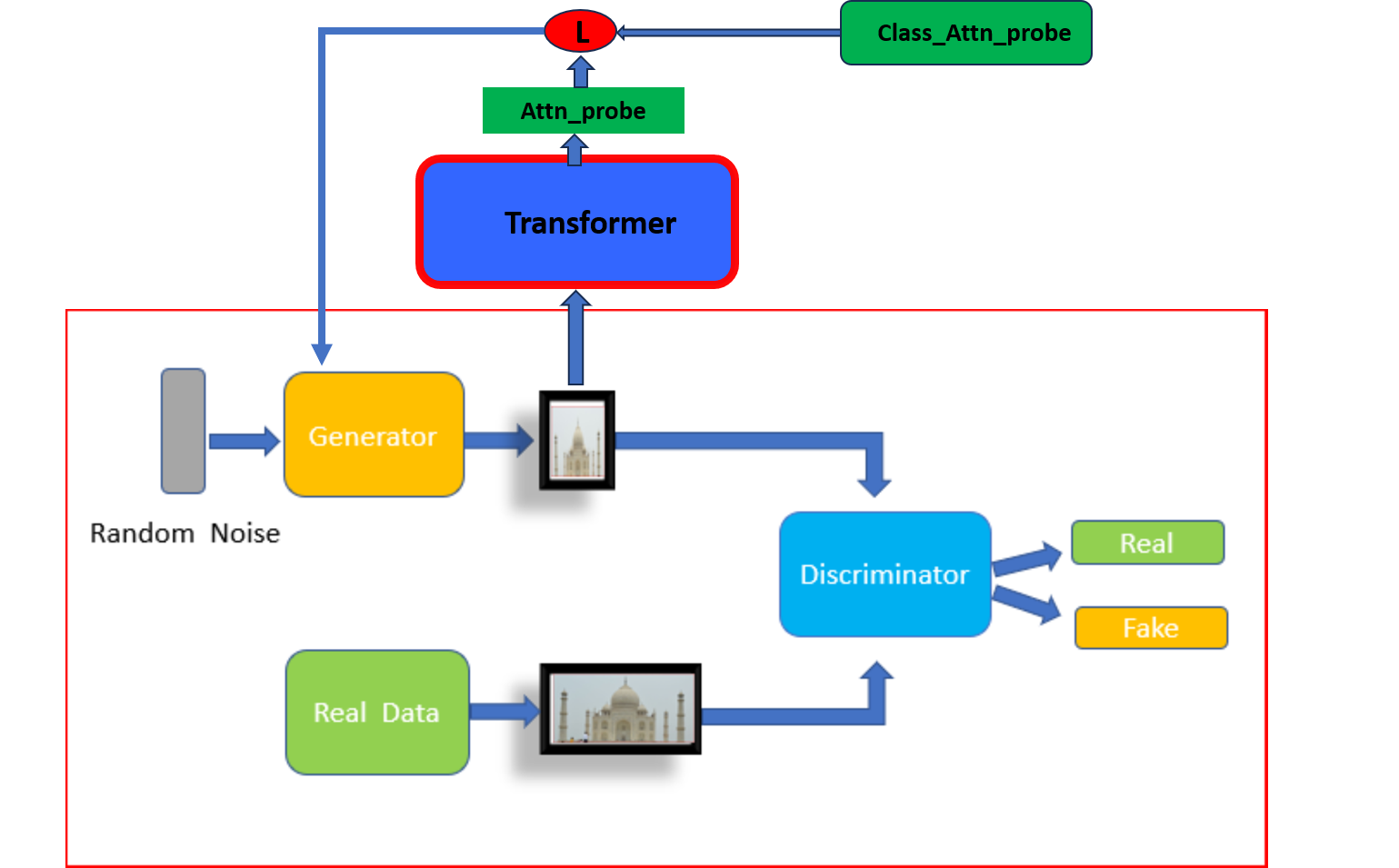} 
  \caption{Proposed Transformer-Augmented GAN Framework }
  \label{fig:5a}
\end{figure}
 \begin{figure}[H]
  \centering
  \fbox{\resizebox{75mm}{!}{\includegraphics{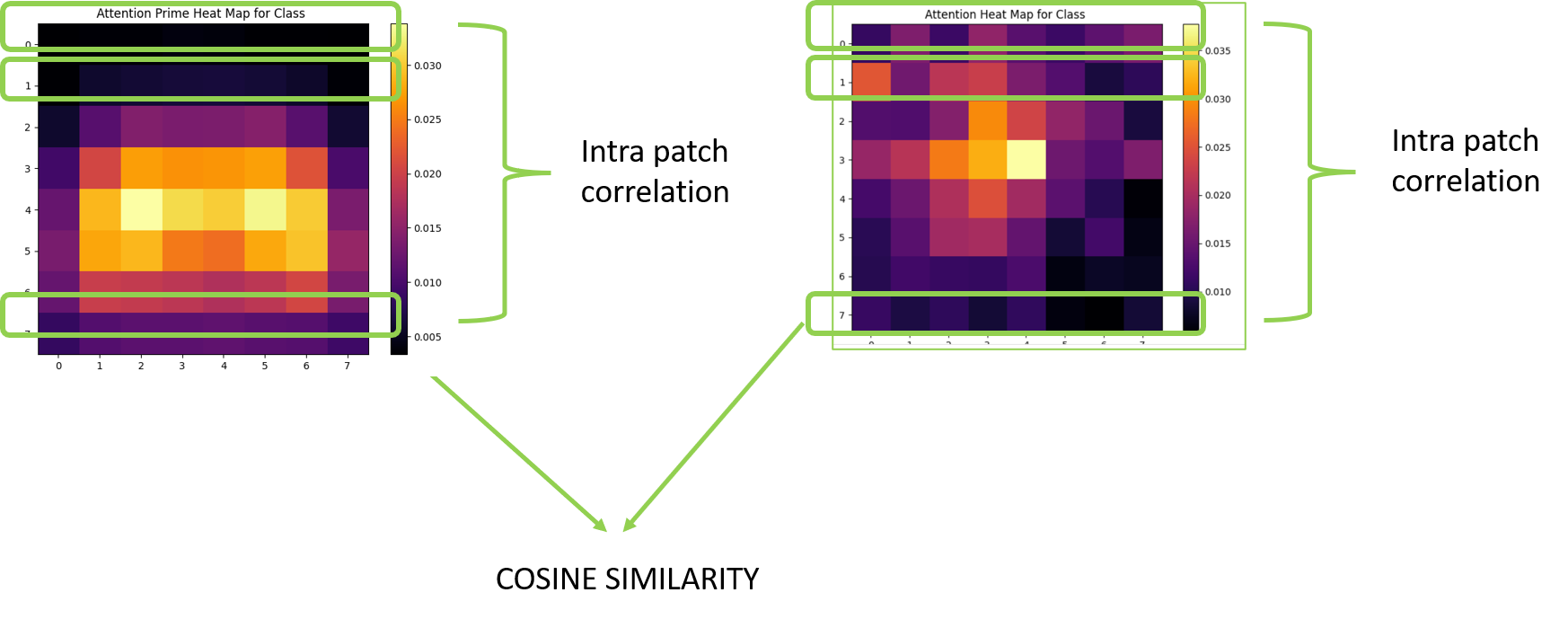}}}
  \caption{Attention loss}
  \label{fig:3}
\end{figure}
First, we extract the features of the real images and the generated images using a pretrained Inception network. The FID score is computed using the Fréchet distance between these two multivariate Gaussian distributions:
\begin{equation}\label{E5}
\text{FID} = \|\mu_r - \mu_g\|^2 + \text{Tr}(\Sigma_r + \Sigma_g - 2(\Sigma_r \Sigma_g)^{1/2})
\end{equation}
where, $\|\mu_r - \mu_g\|^2$ is the squared difference between the means of the real and generated features, and $\text{Tr}$ denotes the trace of a matrix. The term $\text{Tr}(\Sigma_r + \Sigma_g - 2(\Sigma_r \Sigma_g)^{1/2})$ represents the trace of the sum of the covariance matrices of the real and generated features, minus twice the square root of the product of these covariance matrices.\\
The FID score measures the similarity between the real and generated distributions, with lower scores indicating that the generated images are more similar to the real images.\\
For our experiments, the generator architecture includes a label embedding layer that maps class labels to a 5-dimensional space, followed by four transposed convolutional layers. The first transposed convolutional layer expands the input to 512 channels with a kernel size of 4x4 and a stride of 2, followed by batch normalization. This is followed by subsequent layers reducing the number of channels to 256, 128, and finally 3, with batch normalization applied after each layer except the last.\\
The discriminator architecture starts with a label embedding layer mapping class labels to a 1024-dimensional space. This is followed by four convolutional layers. The first layer has 64 channels with a 4x4 kernel and a stride of 2, including dropout. The following layers increase the channels to 128 and 256 with batch normalization and dropout applied after each layer. The final layer outputs a single channel with a 4x4 kernel. Both architectures form the basis for comparing the performance of vanilla GANs and transformer-augmented GANs on the MNIST, CIFAR-10, and CIFAR-100 datasets. The transformer-augmented GANs consistently demonstrated improved FID scores across these datasets, validating our approach.\\
Using this baseline architecture, we conducted experiments with and without transformer augmentation on the MNIST, CIFAR-10, and CIFAR-100 datasets. Our aim in this experiment is not necessarily to produce synthetic images with better FID scores, but rather to demonstrate that for the same number of training epochs, the FID improves when we incorporate the self-attention mechanisms of transformers. For the transformer-augmented GANs, we integrated a trained transformer to compare the cosine similarity of the attention probes of the generated images to the class attention probes. This additional component guides the generator to produce images that align with the typical attention patterns of the target class.
\subsection{Results}

Across all three datasets, the transformer-augmented GANs consistently demonstrated superior performance compared to the vanilla GANs. The FID scores for the transformer-augmented GANs were significantly lower, indicating higher quality and more diverse generated images. This improvement was observed despite using the same number of training epochs for both architectures.\\
For the MNIST dataset, the transformer-augmented GAN achieved a notable reduction in FID score compared to the vanilla GAN. This suggests that incorporating self-attention mechanisms helps the generator focus on critical features, resulting in clearer and more accurate digit representations.
On the CIFAR-10 dataset, the transformer-augmented GAN also outperformed the vanilla GAN, producing images with better FID scores. The self-attention mechanism enabled the model to capture and reproduce intricate patterns and textures more effectively.\\
The CIFAR-100 dataset, which is more complex due to its larger number of classes, further highlighted the benefits of transformer augmentation. The transformer-augmented GAN consistently produced images with improved FID scores, demonstrating the model's ability to generalize well across a diverse set of image classes. 
The hyper parameter we used during our experiment are given in Table 1 as:

\begin{tiny}
    \begin{table}[H]
\centering
\begin{tabular}{|m{1.52cm}|m{1.2cm}|m{.70cm}|m{0.9cm}|m{1.4cm}|}
\hline
\textbf{Hyper-Parameter} &  Learning Rate & Beta & Weight decay & Optimizer \\
\hline
\textbf{Value}& 0.0001& 0.5, 0.999& 2e-5& adam
  \\
\hline
\end{tabular}
\caption{Hyper-Parameter Setting for Proposed Framework}
\end{table}
\end{tiny}
The Table  2 represents tabulated results of our experiment on all three datasets with and without transformer augmentation. 

\begin{figure}[H]
    \centering
    \includegraphics[scale=0.35]{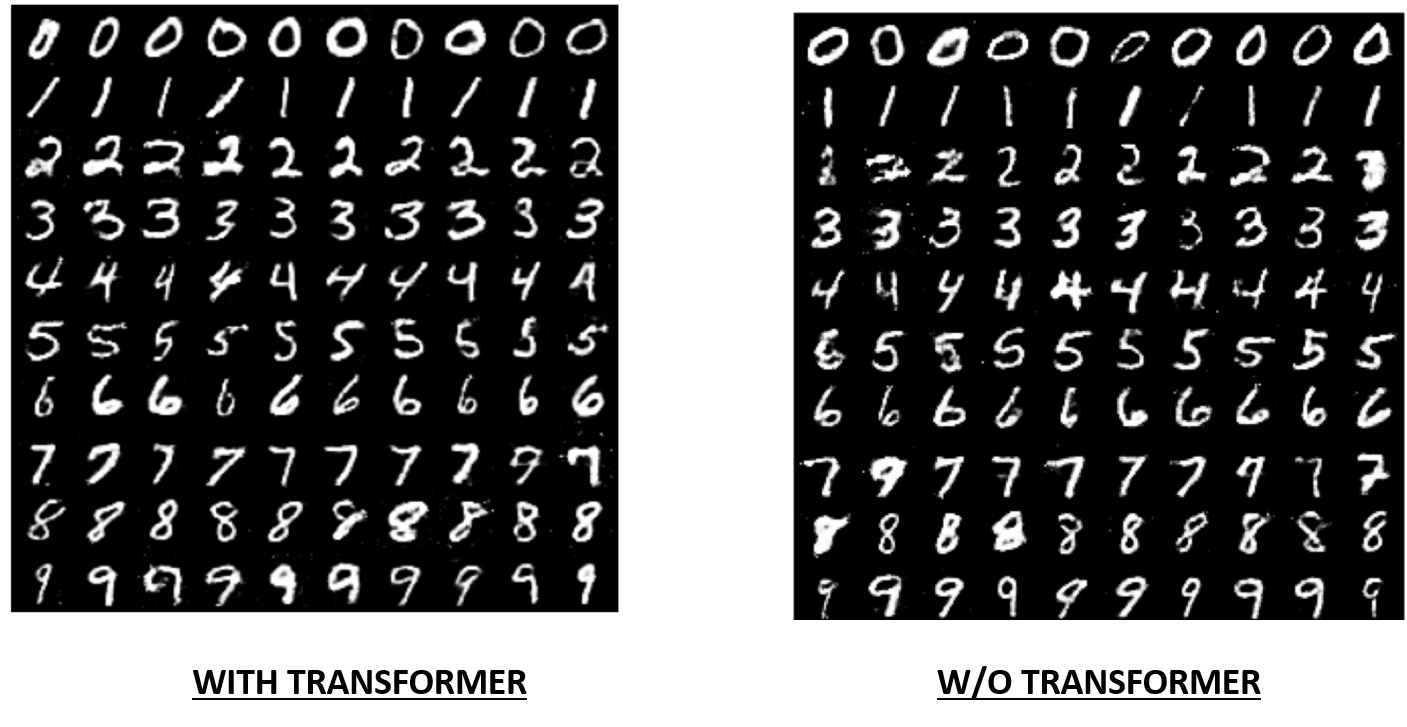}
    \caption{MNIST}
    \label{fig:a1}
\end{figure}

\begin{figure}[H]
    \centering
    \includegraphics[scale=0.35]{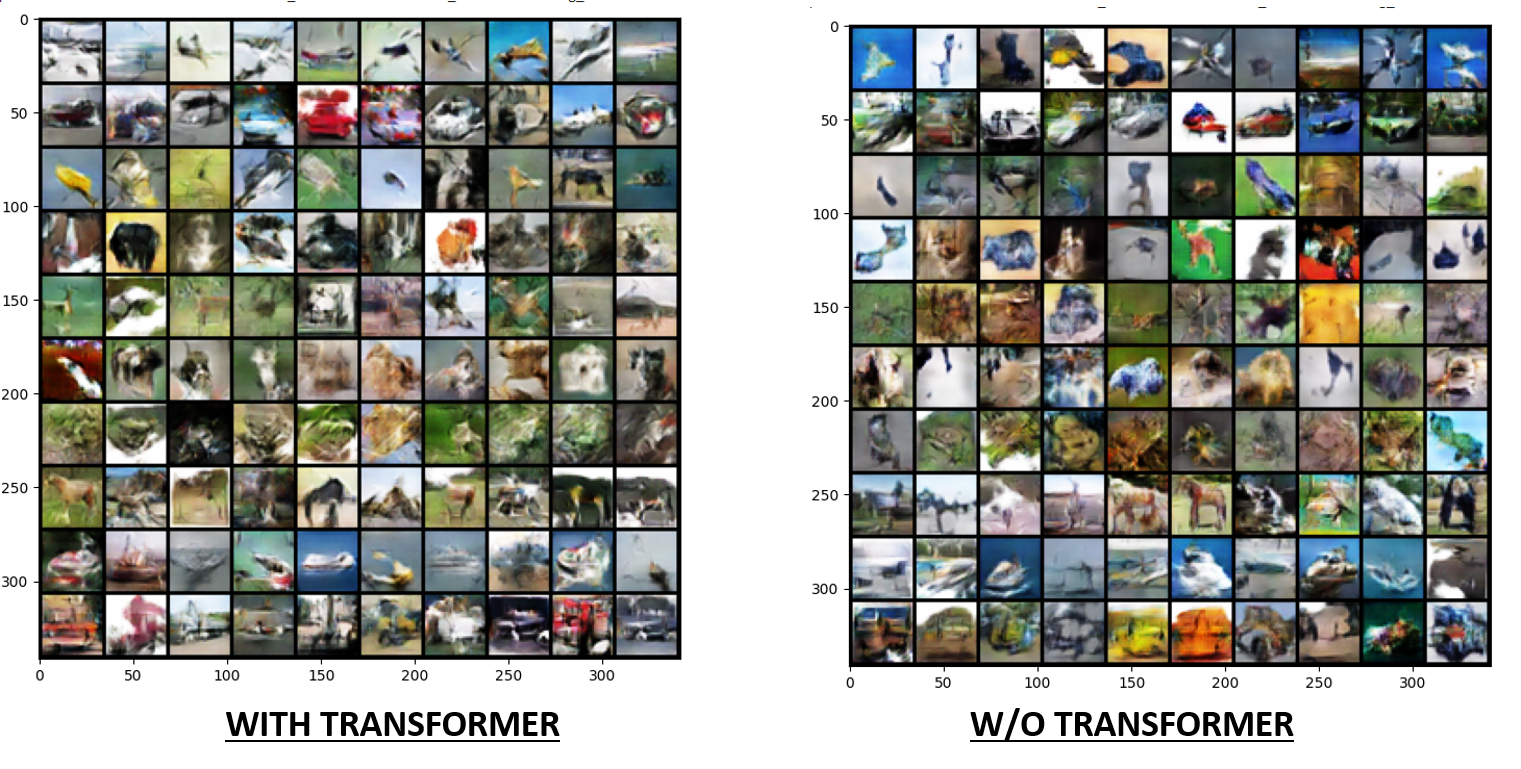}
    \caption{CIFAR10}
    \label{fig:a2}
\end{figure}

\begin{tiny}
\begin{table}[H]
\centering
\begin{tabular}{|m{4.0cm}|m{1.7cm}|m{0.9cm}|}
\hline
\textbf{Data Synthesis Architecture} & \textbf{Dataset} & \textbf{FID Score} \\
\hline
GAN W/o Transformer & MNIST & 27 \\
\hline
GAN Augmented Transformer & MNIST & 23 \\
\hline
GAN W/o Transformer & CIFAR-10 & 76.94 \\
\hline
GAN Augmented Transformer & CIFAR-10 & 70.37 \\
\hline
GAN W/o Transformer & CIFAR-100 & 83.27 \\
\hline
GAN Augmented Transformer & CIFAR-100 & 77.16 \\
\hline
\end{tabular}
\caption{FID Scores with varying architectures}
\end{table}
\end{tiny}
\textbf{Interpretation of Results}
Our hypothesis that incorporating transformer self-attention mechanisms into GANs enhances their performance is substantiated by the consistent improvement in FID scores across different datasets.In order to produce images aligned with the typical attention patterns of the target class, the generator used guidance from the attention probes and the class attention probes. In this integrated approach, issues related to distribution mismatch and bias are addressed, thereby improving the efficiency of knowledge transfer. \\
As a result, the  synthetic images obtained are of greater quality and more contextual accuracy. Our studies' results unequivocally show that transformer-augmented GANs are superior to vanilla GANs in many important ways. The FID scores are improved for the same number of training epochs by utilising transformers' self-attention skills. This shows that transformer-augmented GANs have the ability to produce high-quality synthetic images more quickly, which makes them a viable method for a range of computer vision applications.\\
Figure \ref{fig:a1} and Figure \ref{fig:a2}  shows a comparison of the images produced by the two approaches. The synthetic images produced using transformer augmentation are obviously of a higher calibre and bear a stronger resemblance to the original images found in the dataset. In comparison to the images produced by the vanilla GANs, the transformer-augmented GANs produce images with finer features, higher structural coherence, and fewer arti-facts. Together with the higher FID scores, this visual proof highlights how well self-attention mechanisms work to boost GAN's generating capacities.\\
Our aim in the above section was to improve the quality and diversity of generated images by incorporating transformer self-attention mechanisms into GAN architectures. Our experiments across the MNIST, CIFAR-10, and CIFAR-100 datasets demonstrated that transformer-augmented GANs consistently outperformed vanilla GANs, achieving significantly lower FID scores for the same number of training epochs. \\
This improvement highlights the effectiveness of attention probes and class attention probes in guiding the generator to focus on critical image features. The visual comparison of generated images further validated our findings, showing that transformer augmentation leads to more realistic and detailed synthetic images. These results underscore the potential of transformer-augmented GANs in advancing the field of image generation, paving the way for more efficient and high-quality applications in computer vision.
\section{Data-free Knowledge Distillation- (Classification Task)}

KD is crucial for compressing large, pre-trained models into smaller, more efficient versions while retaining much of the original model's performance. In the context of ViTs, this becomes particularly important given their substantial size and computational demands. However, traditional KD methods rely on access to the original training data, which may not always be feasible due to several reasons.

Given these challenges, there is a critical need to develop data-free distillation methods that can effectively transfer knowledge from a large teacher model to a smaller student model without requiring the original dataset.\\

Mathematically, Data-Free Knowledge Distillation (DFKD) can be formulated as follows:
\begin{equation}
 D=\{X \in \mathbb{R}^{c \times h \times w}, Y = 1, 2, \ldots, K\}
\end{equation}
Where, $Y$ indicates training samples and $K$ represent labels, and \( T(x; \theta_T) \) is a pre-trained teacher network on \( D \).


The main task for student is to minimize the losses that is:
\begin{equation*} 
\min_{\theta_{S}}  \left(\mathcal{L}_{cls} + \mathcal{L}_{KL}\right)
\end{equation*}
In DFKD we learn a lightweight  classification network \( S(x; \theta_S) \) that can imitate the classification capability of \( T(x; \theta_T) \) without using \( D \).
\subsection{Data preprocessing}
In data-free knowledge distillation, data preprocessing involves generating and preparing synthetic data that can effectively mimic the properties of the original training data. This synthetic data serves as a stand-in during the distillation process, enabling the smaller student model to learn from the larger teacher model. \\
In our approach, we utilized Conditional Generative Adversarial Networks (cGANs) to produce synthetic data for the distillation process. The total number of synthetic images generated was 50,000, and these images achieved a Fréchet Inception Distance (FID) score of 12, indicating a high quality of generated data.\\
The steps we followed in data preprocessing for data-free distillation are Random Noise Initialization, Normalization, Scaling and Data Augmentation. By following these preprocessing steps, we ensured that the synthetic data was of high quality and suitable for effective knowledge distillation, thereby facilitating the training of
a lightweight student network that closely mimics the performance of the teacher network without relying on the original training dataset.

\subsection{Experiment setup 2}

In this section, we detail the experimental setup used to perform DFKD for ViTs. The experiments were conducted on two datasets: MNIST and CIFAR-10, using specific teacher and student models designed for each dataset.

For the MNIST dataset, we used a ViT model as the teacher. This model is designed to handle grayscale images with a single channel, and it has an embedding size of 512. The model consists of 3 transformer layers, each with 3 attention heads, and is trained to classify the images into 10 different classes. The total number of parameters in this teacher model is 9,498,122. The student model for MNIST is a smaller, more efficient DeiT model, specifically the DeiT xtiny patch4 28. This model also processes grayscale images and has an embedding size of 128. Similar to the teacher, it has 3 layers but only 2 attention heads per layer, making it significantly lighter with 2,389,514 parameters.

For CIFAR-10, the teacher model is a DeiT base patch4 32, which is tailored for the 32x32 RGB images typical of the CIFAR-10 dataset. This model features an embedding size of 384 and 3 attention heads. It is designed to handle the complexity of CIFAR-10's diverse image set and classify them into 10 classes. For the student model on CIFAR-10, we employed a customized ViT. This model processes 32x32 RGB images with a patch size of 4 and an embedding dimension of 128 and 2 attention heads each. The normalization layer is implemented with LayerNorm, and the total number of parameters is significantly reduced compared to the teacher model. The generic architecture of DFKD is given by the Figure \ref{fig:3} as:
\begin{figure}[H]
  \centering
\includegraphics[scale=0.5]{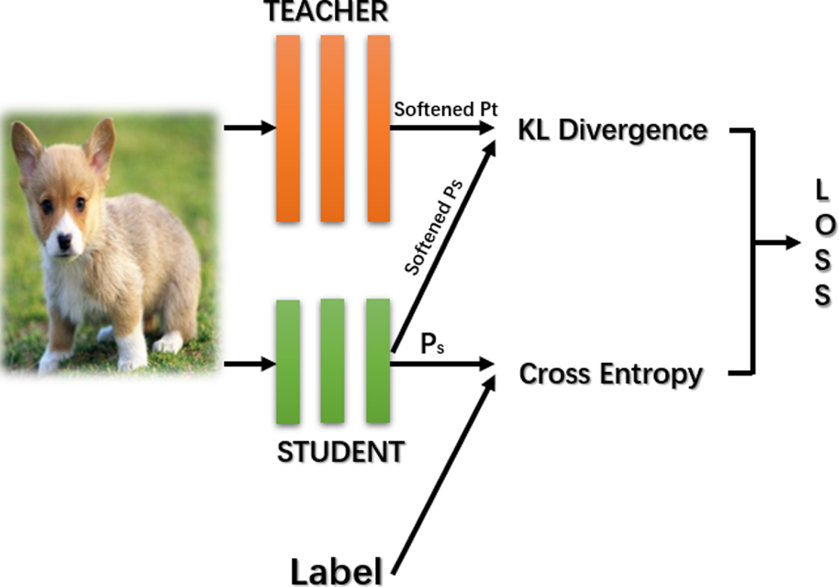}
  \caption{Datafree Knowledge distillation architecture}
  \label{fig:3}
\end{figure}
\subsection{Critical Loss Strategies in DFKD}
In the context of our DFKD experiments for ViT, the primary objective is to transfer the learned knowledge from a large, pre-trained teacher model to a smaller, more efficient student model without relying on the original training data. This process involves the use of multiple loss functions to ensure that the student model effectively mimics the behavior and performance of the teacher model.\\
To achieve effective KD, we employ three distinct loss functions: the knowledge distillation loss $(kd\_loss)$, the cross-entropy loss $(ce\_loss)$, and a custom patch attention probe loss $(patch\_loss)$. Each of these loss functions serves a specific purpose in guiding the training of the student model.\\
\textbf{Knowledge distillation loss $(kd\_loss)$:}
The knowledge distillation loss is calculated by comparing the outputs of the student model with the outputs of the teacher model . This loss ensures that the student model's predictions are aligned with those of the teacher model. It is defined mathematically as the Kullback-Leibler (KL) divergence between the softened predictions of the student and teacher model. 
    \begin{equation}
L_{\text{KD}} = \frac{T^2}{N} \sum_{i=1}^N \sum_{j=1}^C q_{i,j} \log \left( \frac{q_{i,j}}{p_{i,j}} \right)
\end{equation}
where:
\begin{itemize}
    \item \( T \) is the temperature parameter that smooths the probability distributions,
    \item \( p_{i,j} = \frac{\exp(y_{i,j} / T)}{\sum_k \exp(y_{i,k} / T)} \)\\
    is the softened prediction of the student model,
    \item \( q_{i,j} = \frac{\exp(\text{teacher\_scores}_{i,j} / T)}{\sum_k \exp(\text{teacher\_scores}_{i,k} / T)} \)\\
    is the softened prediction of the teacher model,
    \item \( N \) is the number of data points,
    \item \( C \) is the number of classes.
\end{itemize}
\textbf{The cross-entropy loss $(ce\_loss)$:} is a standard classification loss function that measures the difference between the predicted class probabilities and the true class labels. This loss helps the student model to learn to classify the input data correctly.
    \begin{equation}
L_{\text{CE}} = - \frac{1}{N} \sum_{i=1}^N \sum_{j=1}^C y_{i,j} \log(p_{i,j})
\end{equation}
where:
\begin{itemize}
    \item \( y_{i,j} \) is the true label for data point \( i \) and class \( j \),
    \item \( p_{i,j} \) is the predicted probability for data point \( i \) and class \( j \).
\end{itemize}
\textbf{The patch attention probe loss $(patch\_loss)$:} is a custom loss function designed to compare the attention maps of the teacher and student models. This loss encourages the student model to learn similar attention patterns to those of the teacher model, which is crucial for maintaining the model's interpretability and effectiveness.This patch attention probe loss effectively captures the spatial attention patterns of the models, ensuring that the student model learns to focus on similar regions of the input as the teacher model.\\
By combining these loss functions, we create a robust training regimen that enables the student model to effectively learn from the teacher model without access to the original training data, achieving competitive performance while maintaining efficiency. The overall Loss for knowledge distillation is given by :
\begin{equation}
L_{\text{total}} = \lambda_{\text{KD}} L_{\text{KD}} + \lambda_{\text{CE}} L_{\text{CE}} + \lambda_{\text{patch}} L_{\text{patch}}
\end{equation}

where:
\begin{itemize}
    \item \( L_{\text{KD}} \) is the knowledge distillation loss,
    \item \( L_{\text{CE}} \) is the cross-entropy loss,
    \item \( L_{\text{patch}} \) is the patch attention probe loss,
    \item \( \lambda_{\text{KD}} \), \( \lambda_{\text{CE}} \), and \( \lambda_{\text{patch}} \) are the weighting coefficients for each loss component.
\end{itemize}
\subsection{Results} 
    In this section, we present the results of ourknowledge DFKD experiments. Our primary objective was to evaluate the effectiveness of transferring knowledge from a large, pre-trained teacher model to a smaller, more efficient student model without using the original training data. We conducted extensive experiments on multiple datasets, including MNIST and CIFAR-10, to validate our approach. The hyper parameters used by us during the experiment are as under table \ref{T5}. 
    
\begin{tiny}
    \begin{table}[H]
\centering
\begin{tabular}{|m{1.5cm}|m{1.2cm}|m{.70cm}|m{1.1cm}|m{1.37cm}|}
\hline
\textbf{Hyper-Parameter} &  Learning Rate & Beta & Weight-decay & Optimizer \\
\hline
\textbf{Value}& 7.5e-4& 0.5, 0.999& 0.025& AdamW \\
\hline
\end{tabular}
\caption{Hyper-Parameter Values}
\label{T5}
\end{table}
\end{tiny}

\begin{figure}[H]
  \centering
 \includegraphics[scale=0.55]{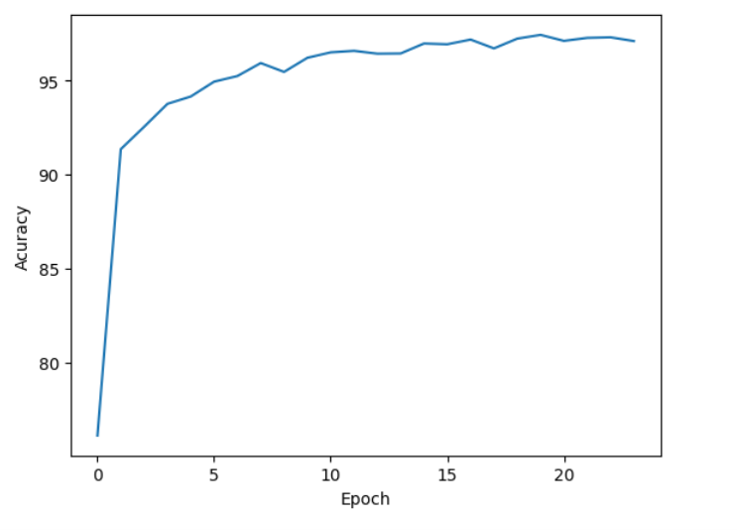} 
  \caption{Student accuracy vs epoch (MNIST)}
  \label{fig:3a}
\end{figure}
\textbf{MNIST Dataset:} For the MNIST dataset, we utilized a ViT as the teacher model and a DeiT-Tiny variant as the student model. The teacher model has a  total 9.4M parameters. The student model, a DeiT\_xtiny\_patch4\_28, amounts to 2.3M parameters. The results showed that our DFKD method successfully distilled the knowledge from the teacher to the student model, achieving competitive performance on the MNIST classification task. The student model demonstrated an accuracy of 97.75\% represented by Table \ref{T-1} and Figure \ref{fig:3a}, which is close to the teacher model's accuracy of 98.32\% without significant  compromise of performance.  The reduction in model size and complexity was significant, making the student model more suitable for deployment on resource-constrained devices in data free environment.\\

\begin{figure}[H]
  \centering
 \includegraphics[scale=.5]{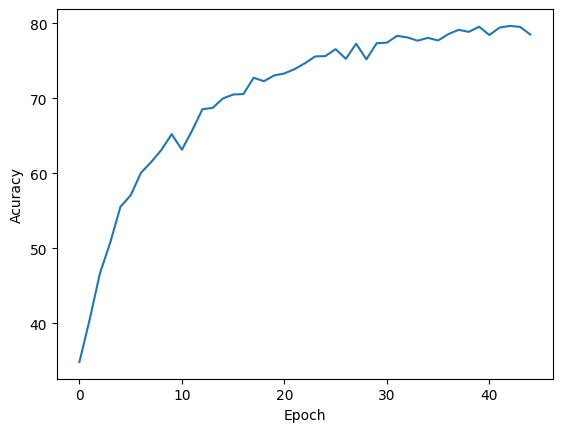} 
  \caption{Accuracy vs Epoch (CIFAR10)}
  \label{fig:11}
\end{figure}

\begin{figure}[H]
  \centering
\includegraphics[scale=0.35]{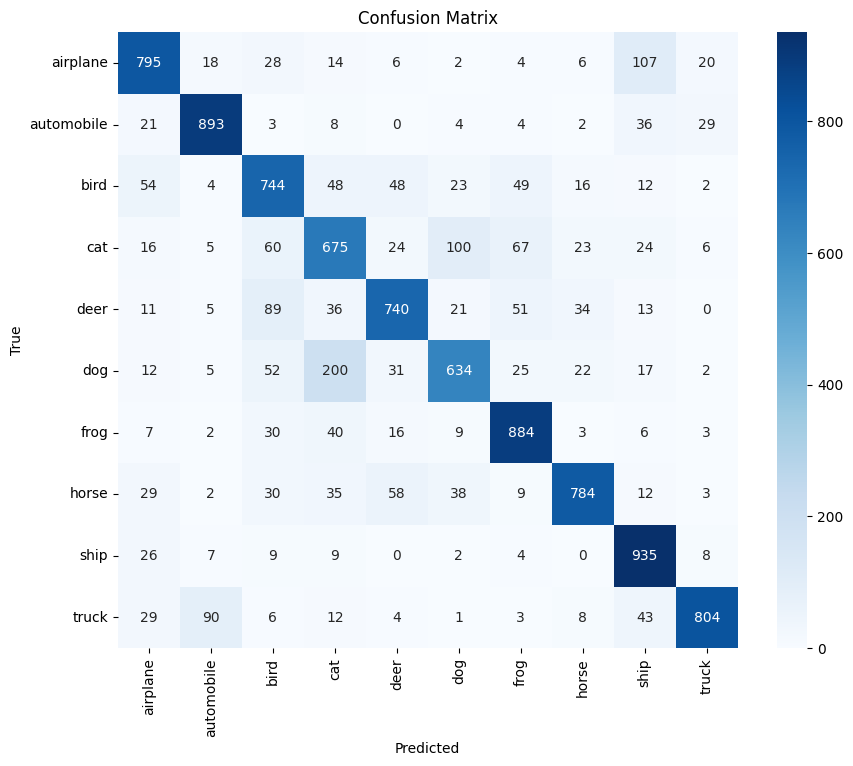} 
  \caption{Confusion matrix student for the final epoch}
  \label{fig:3-a}
\end{figure}

\noindent 
\textbf{CIFAR-10 Dataset:} For the CIFAR-10 dataset, we used a DeiT\_base\_patch4\_32\_teacher as the teacher model and a custom student model based on the ViT model.  The total number of parameters in teacher and student models are 21.3M and 12M respectively.\\
The experimental results on the CIFAR-10 dataset further validated the efficacy of our DFKD approach. The student model achieved an accuracy of 82.3\%, compared to the teacher model's accuracy of 89.5\% on synthetic Data represented by  Table \ref{T-a2} and Figure \ref{fig:11}. This indicates a minor trade-off in performance, which is justified by the substantial reduction in model size and computational requirements.
\begin{tiny}
\begin{table}[H]
\centering
\begin{tabular}{|m{1.4cm}|m{2.5cm}|m{0.8cm}|m{2.5cm}|}
\hline
Type of Data &Teacher (DieT Tiny) Test Acc. & \# Param. & Student (DieT-x Tiny) Test Acc.\\ \hline
True Training  & 98.75\% & 9.4M & 97.75\% \\ \hline
Synthetic & 97.32\% & 2.3M & 96.73\% \\ \hline
\end{tabular}
\caption{KD Performance on MNIST Dataset with and without transformer Augmented GAN}
\label{T-1}
\end{table}
\end{tiny}
\begin{tiny}
\begin{table}[H]
\centering
\begin{tabular}{|m{1.4cm}|m{2.5cm}|m{0.8cm}|m{2.5cm}|}
\hline
Type of Data &Teacher (DieT Tiny) Acc. & \# Param. & Student (DieT-x Tiny) Acc.\\ \hline 
True Training  & 89.51\% & 21.3M & 87.77\% \\ \hline
Synthetic & 84.32\% & 12.0M & 82.37\% \\ \hline
\end{tabular}
\caption{KD Performance on CIFAR-10 Dataset with and without transformer Augmented GAN}
\label{T-a2}
\end{table}
\end{tiny}
Given below is the confusion matrix for the final epoch represented by Figure \ref{fig:3-a}. This analysis provides deeper insights into the model's learning progression and its ability to correctly classify different classes.
\textbf{Analysis of Loss Components}
We performed ablation trials, where we systematically removed one loss component at a time and observed the effect on the performance of the student model, in order to gain a better understanding of the role of each loss component. The findings verified that each loss component—patch attention probe loss, cross-entropy loss, and knowledge distillation loss (KD loss)—has an essential function in the training procedure. More specifically, the teacher and student models' spatial attention patterns aligned better thanks to the patch attention probe loss, which also improved performance and interpretability. The different loss components are indicated by the Figure \ref{fig:3-b} as:
\begin{figure}[H]
    \centering
 \includegraphics[scale=0.55]{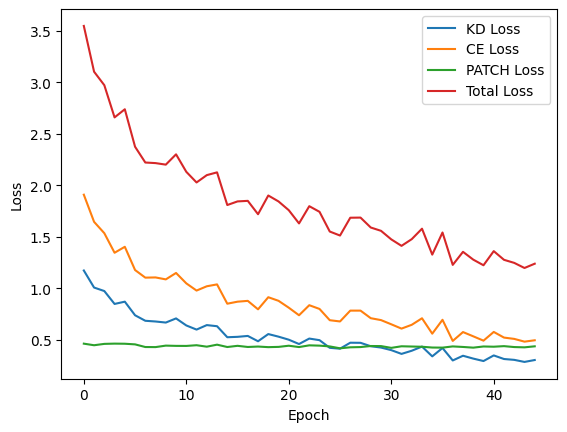} 
    \caption{Loss vs epoch}
    \label{fig:3-b}
\end{figure}
Finally, without relying on the original training data, our DFKD strategy successfully transfers knowledge from big, pre-trained Transformer (teacher) models to smaller, more efficient student models. This method provides an effective way to deploy robust lightweight ViTs in resource-limited settings, ensuring high performance while reducing computational requirements significantly with relying on true distribution of the data.
\section{Data-Free Knowledge Distillation (Detection Task)}
Simple classification is frequently not enough in the majority of real-world applications, particularly crucial ones like military operations, where detection is frequently more important. For activities ranging from surveillance to autonomous navigation and target recognition in defence systems, detection tasks are critical to the identification and localization of objects inside an image. It is therefore crucial to expand our work on KD in ViTs to include detection tasks, given this practical requirement.\\
Classification tasks have been the primary focus of research and optimisation concerning KD. But moving towards detection jobs adds more complexity, including having to anticipate bounding bounds and recognise several items in a single picture. These difficulties call for a more advanced distillation strategy that improves the student model's capacity to precisely detect and localise items in addition to transferring classification capabilities.\\
By creating and utilizing  our proposed DFKD methods especially suited for detection tasks with ViTs, we hope to bridge the gap in this research paper. Also, our aim  is to investigate the specific requirements and techniques associated with transferring information from a teacher model that has already been trained to a lightweight student model, making sure that the student model retains the teacher's detection ability without using the original training set. This method is essential in situations where the usage of real-world datasets is restricted due to transmission, availability, or privacy concerns.\\
By concentrating on detection tasks, we improve the impact and usability of KD in ViTs and bridge the major gap in both commercial and defence applications. The nuances of modifying distillation methods for detection will be covered in detail in this article, along with a thorough framework that upholds data-free limitations and still achieves good performance.\\
DETR, the cutting-edge transformer-based object detection models is used, which has shown impressive results in detection tasks because of its capacity to accurately model links between items and the global context. We intend to condense DETR's complex detection capabilities into a more manageable student model by using it as our teacher model, which will enable deployment in resource-constrained contexts.\\
\subsection{Detection Transformers (DETR):} DETR represent a significant advancement in the field of object detection, leveraging the power of transformer architectures to address the limitations of traditional CNNs. Introduced by Carion et al [41]. in their groundbreaking paper ``End-to-End Object Detection with Transformers'', DETR redefines object detection by directly modeling the global context of an image through self-attention mechanisms.\\
\textbf{Architecture Overview}: 
DETR fundamentally departs from traditional object detection frameworks that rely on region proposal networks (RPNs) or anchor-based methods. Instead, DETR employs a simple and elegant architecture that integrates a transformer encoder-decoder model with a conventional CNN backbone. The architecture can be summarized as follows:
\begin{itemize}
    \item \textbf{Backbone}: A CNN (typically ResNet) extracts feature maps from the input image. These feature maps serve as the input to the transformer.
    \item \textbf{Transformer Encoder}: The feature maps are flattened and embedded, then passed through multiple layers of the transformer encoder. The self-attention mechanism in the encoder enables the model to capture global dependencies across the entire image, ensuring that interactions between all parts of the image are considered.
    \item \textbf{Transformer Decoder}: The decoder takes a fixed number of learned positional embeddings (object queries) and processes them alongside the encoder output. Each object query is responsible for predicting an object in the image. The attention mechanism in the decoder allows each query to focus on relevant parts of the image features produced by the encoder.
    \item \textbf{Prediction Heads}: The output of the decoder is fed into two separate feed-forward networks (FFNs): one for predicting the class labels of the objects and another for predicting the bounding boxes.The generic framework of DETR model is represented by the Figure 14 as:
\end{itemize}
\subsection{Data Preprocessing} For our experiment with DETR, we required a pre-trained DETR model on a detection task. We decided to train the DETR model on a custom drone detection dataset, which included images of four classes: bird, helicopter, drone, and plane. To prepare the dataset for training, we employed several preprocessing steps using the DETR Image Processor from the Hugging Face Transformers library. First, the images were resized so that the shortest edge was 800 pixels and the longest edge was 1333 pixels. This resizing ensured uniformity in image dimensions, facilitating better model training.
\begin{figure}[H]
  \centering
 \includegraphics[scale=0.29]{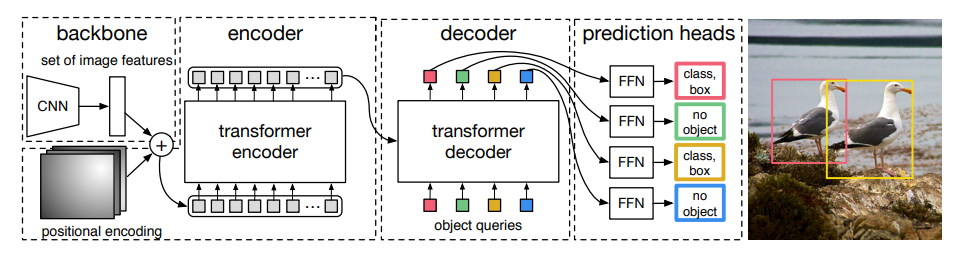}   \caption{DETR architecture}
  \label{fig:3}
\end{figure}
\begin{figure}[H]
    \centering
    \begin{subfigure}{0.45\textwidth}
        \centering
        \includegraphics[width=\linewidth]{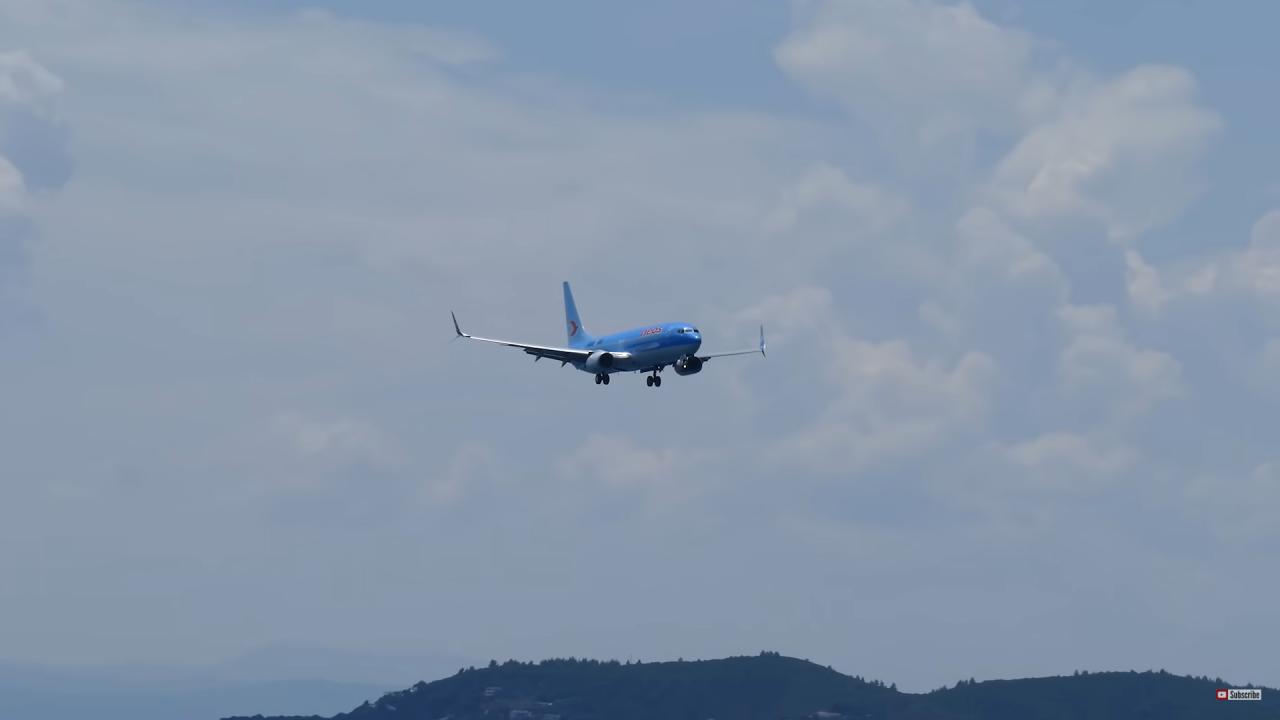}
        \caption{Plane}
        \label{fig:sub1}
    \end{subfigure}\hfill
    \begin{subfigure}{0.45\textwidth}
        \centering
        \includegraphics[width=\linewidth]{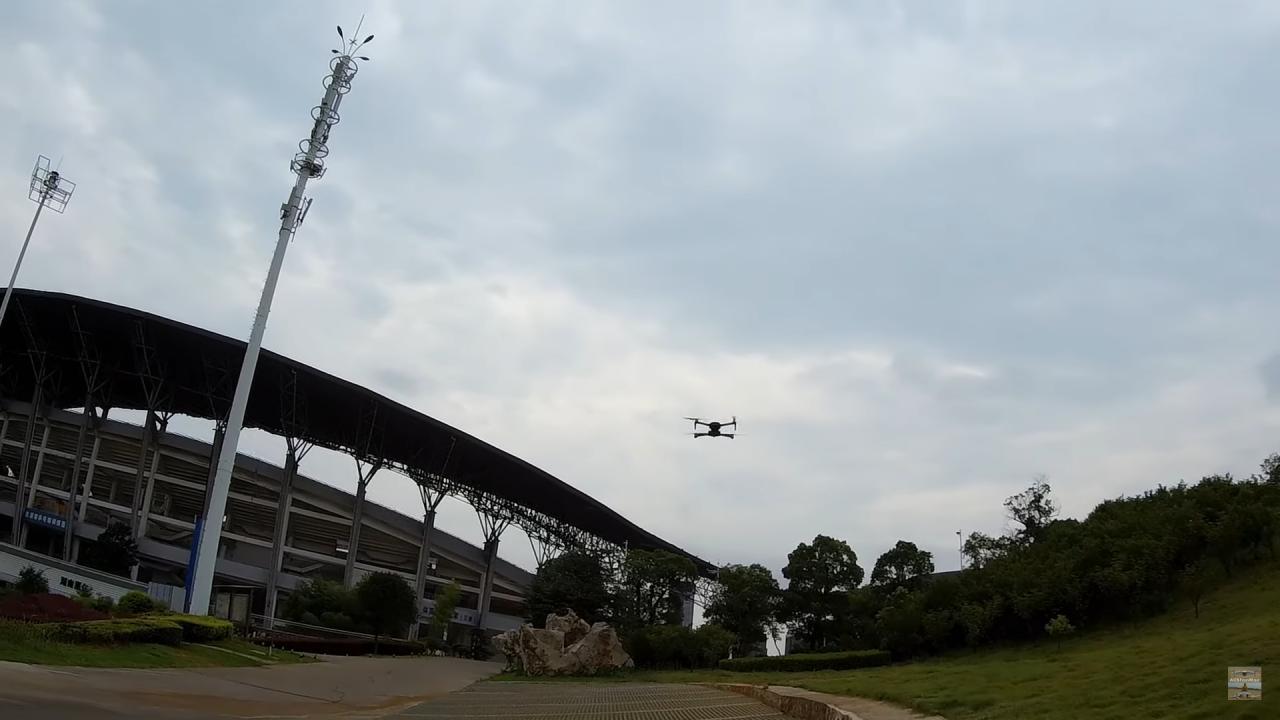}
        \caption{Drone}
        \label{fig:sub2}
    \end{subfigure}
    \vskip\baselineskip
    \begin{subfigure}{0.45\textwidth}
        \centering
        \includegraphics[width=\linewidth]{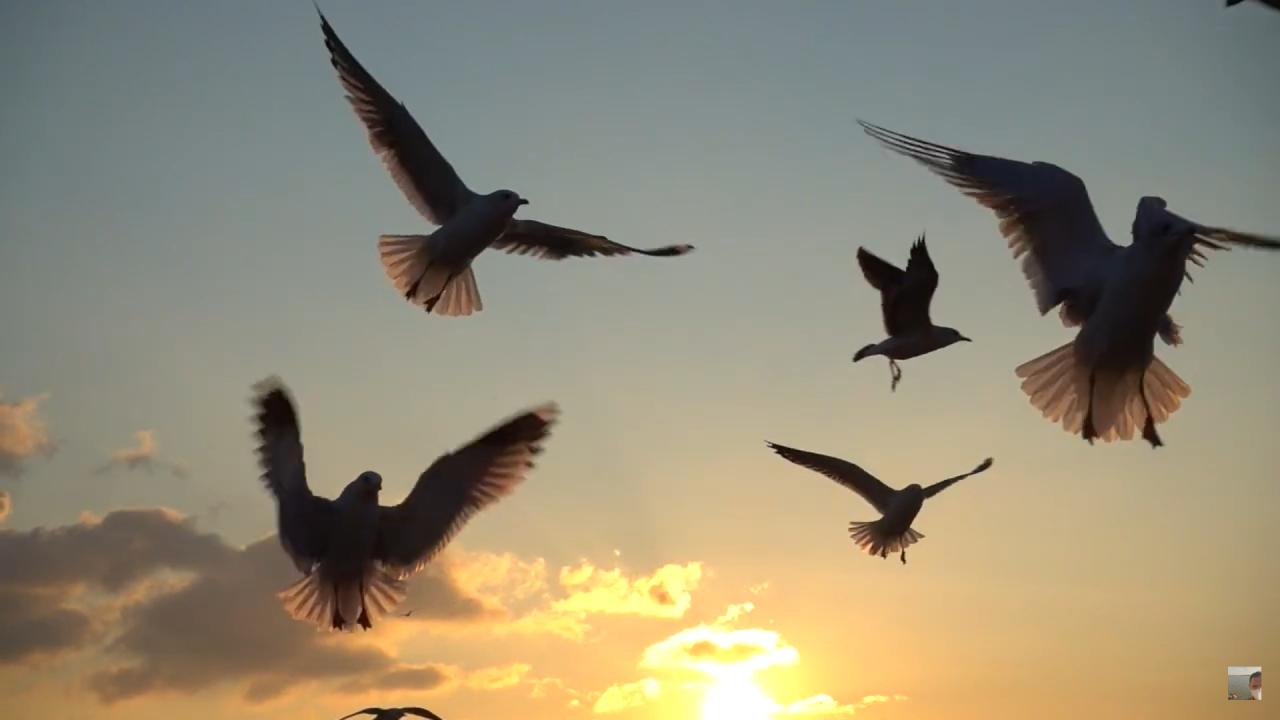}
        \caption{Bird}
        \label{fig:sub3}
    \end{subfigure}\hfill
    \begin{subfigure}{0.45\textwidth}
        \centering
        \includegraphics[width=\linewidth]{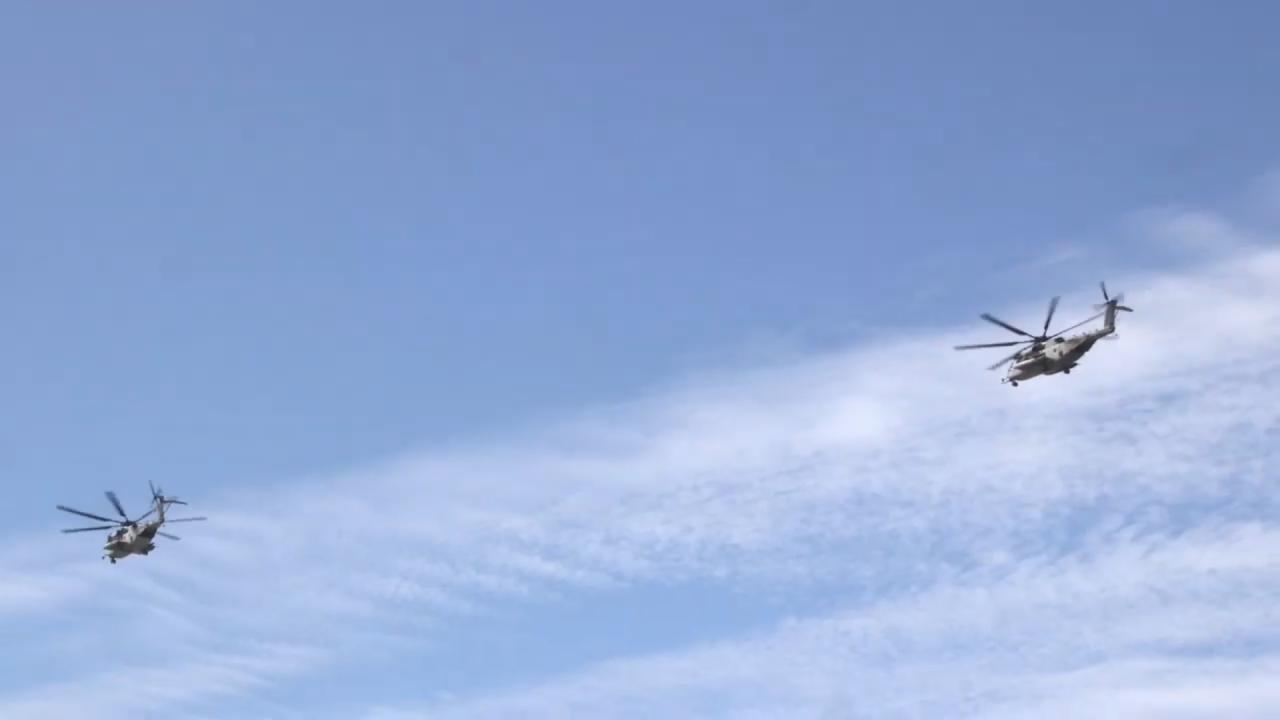}
        \caption{Helicopter}
        \label{fig:sub4}
    \end{subfigure}
    \caption{Sample images of Dataset}
    \label{fig:all_images}    
\end{figure} 
We normalized the images using the mean values (0.485, 0.456, 0.406) and standard deviation values (0.229, 0.224, 0.225) for each color channel. This normalization was essentiasl to adjust the pixel values, making the training process more stable and efficient. Annotations were converted to the COCO [42] format required by DETR, ensuring compatibility and ease of processing. Padding was applied to the images to meet the required size specifications without distorting the image content. Additionally, pixel values were rescaled by a factor of 0.00392156862745098 to standardize the input data.

Using the Coco-Detection class, we loaded the dataset and applied these preprocessing steps, ensuring each image and its corresponding annotations were correctly formatted.\\
Next, we utilized a DETR-augmented GAN in similar fashion as illustrated by Figure 5 to create synthetic images close to real-world data distributions. The DETR-GAN model combines the strengths of DETRs for accurate object detection with the generative capabilities of GANs for creating synthetic data of high quality. This comprehensive preprocessing pipeline was crucial for training the DETR model effectively on our drone detection dataset, enabling high-quality object detection leveraging data free environment in our experiments. 
\section{Experiment Setup}
In this section, we outlined the experiment setup for the DFKD technique, employing a Teacher model based on DETR with specific configurations, and a Student model also based on DETR but with reduced complexity. The goal is to distill knowledge from the Teacher to the Student model without relying on data. 
\subsection{Teacher Model Configuration (DETR)} 
\begin{figure}[H]
  \centering
  \includegraphics[scale=1]{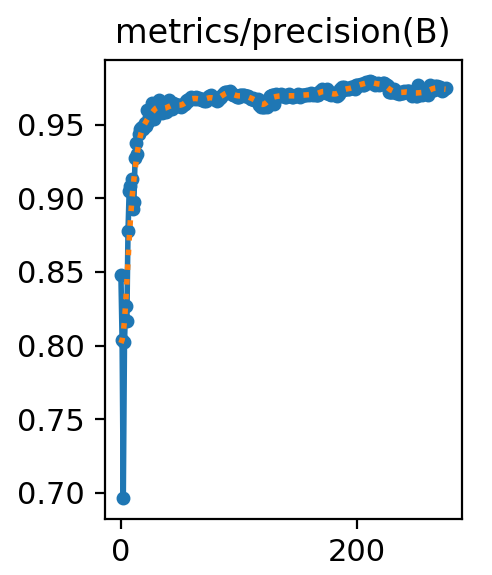} 
  \caption{Precision metrics}
  \label{fig:3-c}
\end{figure}

\begin{figure}[H]
  \centering
\includegraphics[scale=0.65]{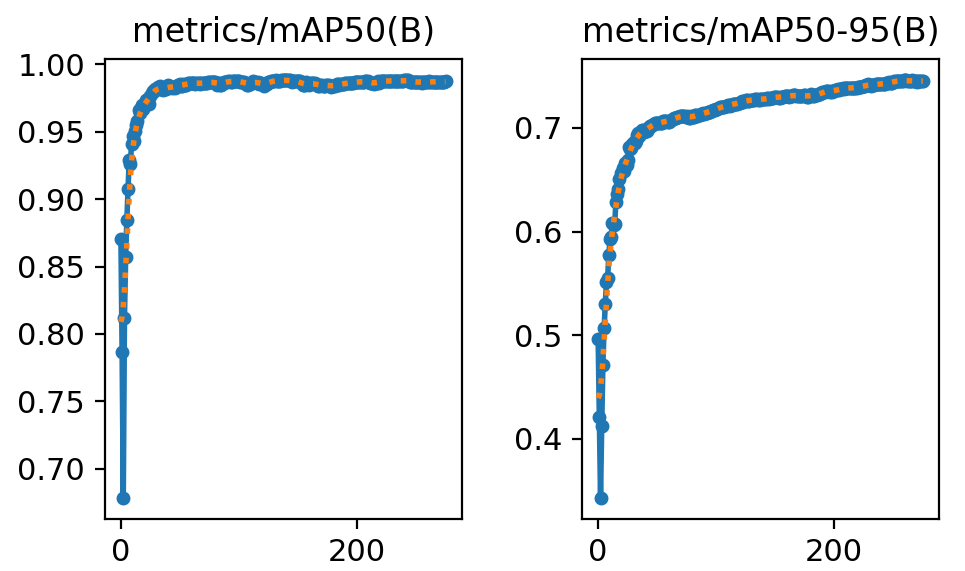}
  \caption{mAP metrics}
  \label{fig:3-d}
\end{figure}

The Teacher model utilized in this experiment is based on the DETR architecture, specifically employing the ResNet-50 backbone with the following specifications:
\begin{itemize}
    \item Backbone: ResNet-50
    \item Encoder and Decoder Layers: 6
    \item Encoder and Decoder Attention Heads: 8
    \item Trainable Parameters: 41.3 million
    \item Non-trainable Parameters:  222 thousand
    \item Total Number of Parameters:41.5 million
    \item Total Estimated Model Parameter Size: 166.008 MB
\end{itemize}
The DETR model serves as the knowledge source, possessing a substantial number of parameters and complex attention mechanisms to accurately perform object detection tasks. Its rich representation is to be distilled into a smaller, more lightweight Student model. As we were dealing with the custom dataset we trained the teacher model from scratch. The graphs below (Figure \ref{fig:3-c} and Figure \ref{fig:3-d}) shows the precision and mAP over the training cycle of teacher model.

\subsection{Student Model Configuration:} The Student model is a simplified version of the DETR architecture, designed for efficient inference and reduced computational cost. Its specifications are as follows:
\begin{itemize}
    \item Backbone: ResNet-18
    \item Encoder and Decoder Layers: 2
    \item Total Number of Parameters: 1.7 million
\end{itemize}
Compared to the Teacher model, the Student model has a significantly reduced parameter count and architectural complexity. By employing a shallower backbone and fewer layers, it aims to strike a balance between computational efficiency and performance.\\
\textbf{Experimental Rationale:} The choice of DETR as the base architecture for both Teacher and Student models ensures consistency in representation learning and knowledge transfer. By varying the depth and complexity of the models, we aim to observe the impact of model architecture on distillation performance.

The utilization of ResNet backbones in both models enables feature extraction from input images, crucial for object detection tasks. However, the stark contrast in parameter count and architectural complexity between the Teacher and Student models introduces a challenging scenario for knowledge distillation.

In the subsequent sections, we delve into the methodology employed for distilling knowledge from the Teacher to the Student model, leveraging data-free techniques to facilitate efficient knowledge transfer and model compression.
\section{Knowledge Distillation} We go into the specifics of the loss functions that are used to extract knowledge from the Teacher to the Student model in this section. In order to achieve equivalent performance with less computational expense, the Student model is trained to replicate the behaviour and predictions of the more sophisticated Teacher model.\\
The following loss functions are employed in the KD process:\\

\textbf{1. Classification Loss  }:
\begin{small}
    \begin{equation}
        \mathcal{L}_{\text{classification}} =\frac{1}{N} \sum_{i=1}^{N} \left[(y_i-1) \log(1 - \sigma(x_i))-y_i \log(\sigma(x_i))  \right]
    \end{equation}
\end{small}
    where: 
    \begin{itemize}
        \item  $  \sigma(x) = \frac{1}{1 + e^{-x}} $
        \item \( N \) is the number of samples.
        \item \( y_i \) represents the expanded targets (ground truth).
        \item \( x_i \) represents the student logits (predicted logits).
        \item \( \sigma(x_i) \) is the sigmoid function applied to the logits.
    \end{itemize}
\textbf{Description}: The classification loss measures the discrepancy between the predicted class probabilities by the Student model   (\( \text{student\_logits} \))  and the ground truth class labels (\( \text{expanded\_targets} \)). It employs Binary Cross-Entropy (BCE) with Logits Loss to compute the classification error.\\ 

\textbf{2. Bounding Box Loss}:
 \begin{small}
        \begin{equation}
        \mathcal{L}_{\text{bbox}} =\mathcal{L}_{\text{SmoothL1Loss}}(A,B)
    \end{equation}
Where, $A=\text{student\_pred\_boxes} \text{ and } B=\text{expanded\_boxes}$\\
 \end{small}
 The mathematical formula for the Smooth L1 Loss is:
\begin{small}
        \begin{equation}
        \mathcal{L}_{\text{SmoothL1}}(x, y) = 
        \begin{cases} 
            0.5 (x - y)^2 & \text{if } |x - y| < 1 \\
            |x - y| - 0.5 & \text{otherwise}
        \end{cases}
    \end{equation}
\end{small}
    where:
    \begin{itemize}
        \item \( \mathcal{L}_{\text{bbox}} \) is the bounding box loss.
        \item \( x \) represents the predicted bounding box coordinates (student\_pred\_boxes).
        \item \( y \) represents the ground truth bounding box coordinates (expanded\_boxes).
    \end{itemize} 
\textbf{Description}: The bounding box loss quantifies the difference between the predicted bounding box coordinates (\( \text{student\_pred\_boxes} \)) by the Student model and the ground truth bounding box coordinates (\( \text{expanded\_boxes} \)). It utilizes Smooth L1 Loss to calculate the regression error.\\

\textbf{3. Distillation Loss}:
\begin{small}
        \begin{equation}
        \mathcal{L}_{\text{distill}} = \text{KLDivLoss} \left( X, Y \right)
    \end{equation}
\end{small}
\begin{tiny}
        $X=\log{\- softmax} \left( \frac{\text{student\_logits}}{\text{temperature}} \right) \text{ and } Y=\text{softmax} \left( \frac{\text{teacher\_logits}}{\text{temperature}} \right)$
\end{tiny}

\textbf{Description}: The  final distillation loss measures the disparity between the softened predictions of the Student model 
    \(\log\_softmax \left( \frac{\text{student\_logits}}{\text{temperature}} \right)\) and the Teacher model 
    \(\text{softmax} \left( \frac{\text{teacher\_logits}}{\text{temperature}} \right)\). 
    It employs Kullback-Leibler (KL) Divergence Loss to compute the divergence between probability distributions.
    
    The expanded mathematical formula for the KL Divergence Loss is:
    \begin{equation*}
        \mathcal{L}_{\text{KL}}(P, Q) = \sum_{i} P(i) \log \left( \frac{P(i)}{Q(i)} \right)
    \end{equation*}
   \begin{itemize}
        \item \( P \) is the softened predictions from the student model: \( \log\_softmax \left( \frac{\text{student\_logits}}{\text{temperature}} \right) \)
        \item \( Q \) is the softened predictions from the teacher model: \( \text{softmax} \left( \frac{\text{teacher\_logits}}{\text{temperature}} \right) \)
    \end{itemize} 
The Student model learns to minimize the difference between their softer predictions while reproducing the Teacher model's classification and regression outputs by mixing these loss functions. Efficient knowledge transfer from the Teacher to the Student model is made possible by the distillation process, which allows the Student model to attain comparable performance with less computational complexity.

These loss functions are iteratively calculated for every batch of data in the training loop, and the gradients are back-propagated via the Student model to update its parameters. In a similar vein, the performance of the simplified Student model is appraised through the evaluation of the loss metrics during validation.
\section{Results}
\begin{figure}[H]
  \centering
\includegraphics[scale=0.25]{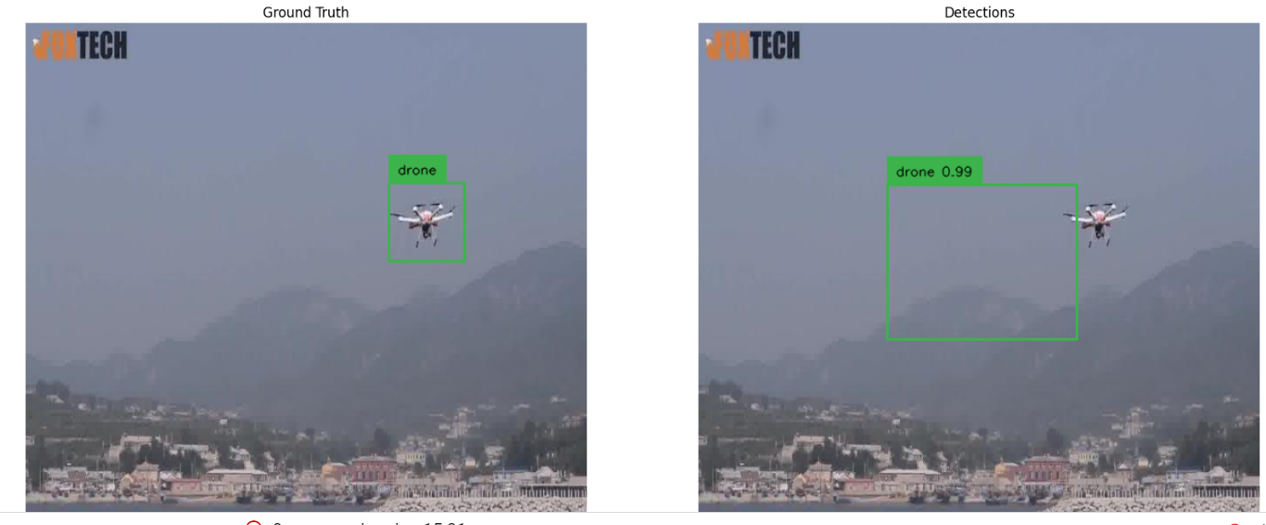}
  \caption{Sample detection by Student model}
  \label{fig:4-a}
\end{figure}

\begin{figure}[H]
  \centering 
  \includegraphics[scale=0.40]{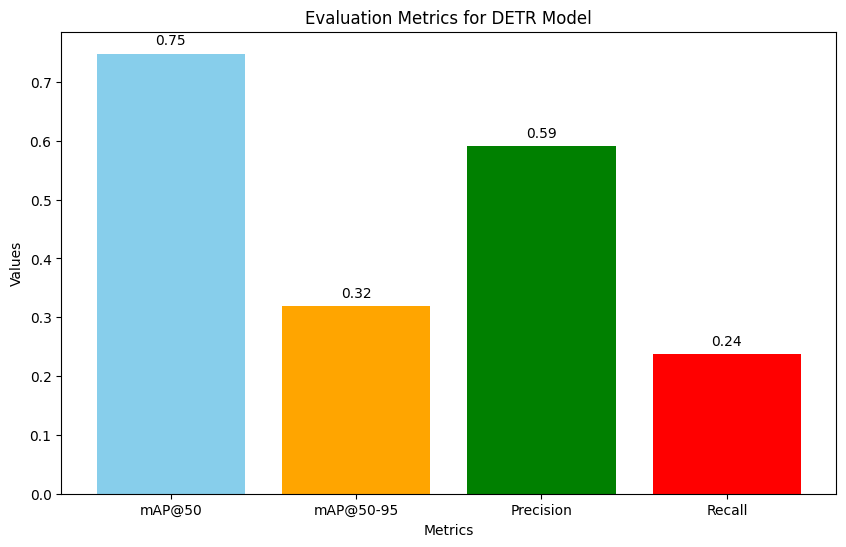}
  \caption{mAP metrics- Student model}
  \label{fig:3-e}
\end{figure}

To confirm the efficacy of the suggested DFKD technique, a thorough evaluation of its performance on detection tasks was conducted. The trials validated the potential of our technique in practical settings by demonstrating notable enhancements in student model performance and efficiency  Figure \ref{fig:4-a}.\\
\textbf{Performance Metrics:} To evaluate the effectiveness of the DFKD technique, we employed the following performance metrics.\\
\textbf{Mean Average Precision (mAP):} This metric evaluates the accuracy of object detection, accounting for both precision and recall  Figure \ref{fig:3-e}.
In summary, our DFKD method for detection tasks has shown to be quite successful, yielding notable gains in efficiency and accuracy. The successful deployment of the lightweight DETR model on devices with limited resources highlights the usefulness of our methodology. Our approach provides a scalable and adaptable solution for a range of applications, including UAV identification, and opens the door for future developments in real-time object detection.
\section{Conclusion}
Transformers have proven to be quite effective in a variety of vision tasks; yet, their size and complexity frequently make it difficult to use them in practical applications. Our research was driven by this challenge to create methods for DFKD in vision transformers, which tackle the two problems of big transformer size and lack of data for KD.

To exploit the power of transformers in picture production, we modified conventional Generative Adversarial Networks (GANs) using a revolutionary technique that we developed in our approach. To be more precise, we implemented attention techniques at the patch level, which greatly increased GAN effectiveness and efficiency. In doing so, we were able to preserve a lightweight and deployable model while taking advantage of transformers' enhanced picture creation capabilities.

\end{document}